\documentclass[superscriptaddress, nofootinbib, longbibliography, amsmath, amssymb, twocolumn]{revtex4-1}
\usepackage[margin=1in]{geometry} 
\usepackage[english]{babel}
\usepackage[utf8]{inputenc}

\usepackage{graphicx}
\usepackage{xspace}
\usepackage{bm}
\usepackage{siunitx}
\DeclareSIUnit\angstrom{\text {Å}}
\usepackage{hyperref}
\DeclareRobustCommand{\gobblefive}[5]{}
\newcommand*{\SkipTocEntry}{\addtocontents{toc}{\gobblefive}}

\hypersetup{hidelinks,colorlinks=true,breaklinks=true,urlcolor=blue,allcolors=blue,pdftitle={Title},pdfauthor={Author}}

\usepackage{listings, xcolor}
\definecolor{codegreen}{rgb}{0,0.6,0}
\definecolor{codegray}{rgb}{0.5,0.5,0.5}
\definecolor{codepurple}{rgb}{0.58,0,0.82}
\definecolor{tqblue}{HTML}{08293d}
\definecolor{backcolour}{HTML}{fefdf5}

\lstdefinestyle{pythonstyle}{
    backgroundcolor=\color{backcolour},   
    commentstyle=\color{codegreen},
    keywordstyle=\color{magenta},
    numberstyle=\tiny\color{codegray},
    stringstyle=\color{codepurple},
    basicstyle=\ttfamily\footnotesize\color{tqblue},
    breakatwhitespace=false,         
    breaklines=true,
    postbreak=\mbox{\textcolor{magenta}{$\hookrightarrow$}\space},                 
    captionpos=b,                    
    keepspaces=true,                 
    numbers=left,                    
    numbersep=5pt,                  
    showspaces=false,                
    showstringspaces=false,
    showtabs=false,                  
    tabsize=2
}

\UseRawInputEncoding 
\begin{document}

\title{Predicting the Future of AI with AI:\\High-Quality link prediction in an exponentially growing knowledge network}

\author{Mario~Krenn~}
\email{mario.krenn@mpl.mpg.de}
\affiliation{Max Planck Institute for the Science of Light (MPL), Erlangen, Germany.}

\author{Lorenzo Buffoni~}
\affiliation{Instituto de Telecomunica\c c\~oes, Lisbon, Portugal.}

\author{Bruno Coutinho~}
\affiliation{Instituto de Telecomunica\c c\~oes, Lisbon, Portugal.}

\author{Sagi~Eppel}
\affiliation{University of Toronto, Canada.}

\author{Jacob Gates Foster}
\affiliation{University of California Los Angeles, USA.}

\author{Andrew~Gritsevskiy~}
\affiliation{University of Toronto, Canada.}
\affiliation{Cavendish Laboratories, Cavendish, Vermont, USA.}
\affiliation{Institute of Advanced Research in Artificial Intelligence (IARAI), Vienna, Austria.}

\author{Harlin~Lee~}
\affiliation{University of California Los Angeles, USA.}

\author{Yichao Lu}
\affiliation{Layer 6 AI, Toronto, Canada.}

\author{Jo\~{a}o P. Moutinho~}
\affiliation{Instituto de Telecomunica\c c\~oes, Lisbon, Portugal.}

\author{Nima~Sanjabi}
\affiliation{Independent Researcher, Barcelona, Spain.}

\author{Rishi~Sonthalia~}
\affiliation{University of California Los Angeles, USA.}

\author{Ngoc Mai Tran}
\affiliation{University of Texas at Austin, USA.}

\author{Francisco Valente~}
\affiliation{Independent Researcher, Leiria, Portugal.}

\author{Yangxinyu Xie}
\affiliation{University of Pennsylvania, USA.}

\author{Rose Yu~}
\affiliation{University of California, San Diego, USA.}

\author{Michael Kopp~}
\affiliation{Institute of Advanced Research in Artificial Intelligence (IARAI), Vienna, Austria.}
\begin{abstract}
A tool that could suggest new personalized research directions and ideas by taking insights from the scientific literature could significantly accelerate the progress of science. A field that might benefit from such an approach is artificial intelligence (AI) research, where the number of scientific publications has been growing exponentially over the last years, making it challenging for human researchers to keep track of the progress. Here, we use AI techniques to predict the future research directions of AI itself. We develop a new graph-based benchmark based on real-world data -- the \texttt{Science4Cast} benchmark, which aims to predict the future state of an evolving semantic network of AI. For that, we use more than 100,000 research papers and build up a knowledge network with more than 64,000 concept nodes. We then present ten diverse methods to tackle this task, ranging from pure statistical to pure learning methods. Surprisingly, the most powerful methods use a carefully curated set of network features, rather than an end-to-end AI approach. It indicates a great potential that can be unleashed for purely ML approaches without human knowledge. Ultimately, better predictions of new future research directions will be a crucial component of more advanced research suggestion tools.
\end{abstract}

\maketitle

\newpage

\begin{figure}[ht]
\centering
\includegraphics[width=0.45\textwidth]{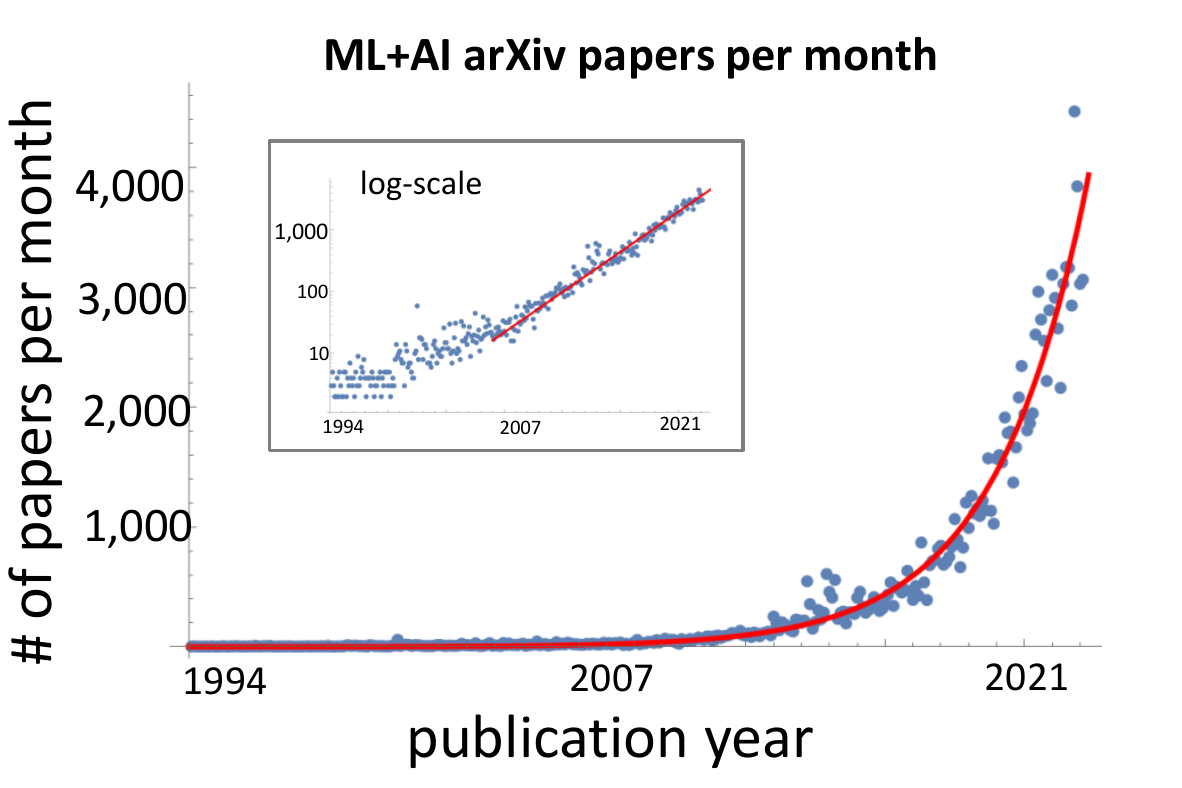}
\caption{\textbf{Number of papers published per months in the arXiv categories of AI grow exponentially.} The doubling rate of papers per months is roughly 23 months, which might lead to problems for publishing in these fields, at some point. The categories are \texttt{cs.AI}, \texttt{cs.LG}, \texttt{cs.NE}, and \texttt{stat.ML}.}
\label{fig:numOfPapersAI}
\end{figure}

\section{Introduction and Motivation}
The corpus of scientific literature grows at an ever-increasing speed. Specifically, in the field of Artificial Intelligence (AI) and Machine Learning (ML), the number of papers every month grows exponentially with a doubling rate of roughly 23 months (see Fig. \ref{fig:numOfPapersAI}). Simultaneously, the AI community is embracing diverse ideas from many disciplines such as mathematics, statistics, and physics, making it challenging to organize different ideas and uncover new scientific connections. We envision a computer program that can automatically read, comprehend and act on AI literature. It can predict and suggest meaningful research ideas that transcend individual knowledge and cross-domain boundaries. If successful, it could significantly improve the productivity of AI researchers, open up new avenues of research, and help drive progress in the field.

Here, we address this important and challenging vision. New research ideas often result from drawing novel connections between seemingly unrelated concepts \cite{evans2011metaknowledge, fortunato2018science, wang2021science}. Therefore, we formulate the evolution of AI literature as a temporal network modelling task. We created an evolving semantic network characterizing the content and evolution of the scientific literature in the field of AI since 1994. The network contains about 64,000 nodes (each representing a concept used in an AI paper) and 18 million edges that connect two concepts when they were investigated jointly in a scientific paper. 

We use the semantic network as an input to 10 diverse statistical and machine-learning methods to predict the future evolution of the semantic network with high accuracy. That is, we can predict which combinations of concepts AI researchers will investigate in the future. Being able to predict what scientists will work on is a first crucial step for \textit{suggesting} new topics that might have a high impact.

Several of the methods presented in this paper have been contributions to the \texttt{Science4Cast} competition hosted by \textit{IEEE BigData 2021}, which ran from August to November 2021. Broadly, we can divide the methods into two classes: methods that use hand-crafted network-theoretical features, and those that automatically learn features. We found that models using carefully hand-crafted features outperform methods that attempt to learn features autonomously. This (somewhat surprising) finding indicates a great potential for improvements of models free of human priors.

Our manuscript has several purposes. First, we introduce a new meaningful benchmark for AI on real-world graphs. Second, we provide nearly 10 diverse methods that solve this benchmark. Third, we explain how solving this task could become an essential ingredient for the big picture goal of having a tool that could suggest meaningful research directions for scientists in AI or in other disciplines.\footnote{\href{https://github.com/artificial-scientist-lab/FutureOfAIviaAI}{github.com/artificial-scientist-lab/FutureOfAIviaAI}}

The manuscript is structured in the following way. We first introduce more background into semantic networks and how they can help to suggest new ideas. Then we explain how we generate the dataset and some of its network-theoretical properties. Then we briefly explain the 10 methods that we have investigated to solve the task. We conclude with a number of important open questions that could bring us further toward the goal of AI-based suggestions for research directions.

\begin{figure*}[htbp!]
\centering
\includegraphics[width=0.95\textwidth]{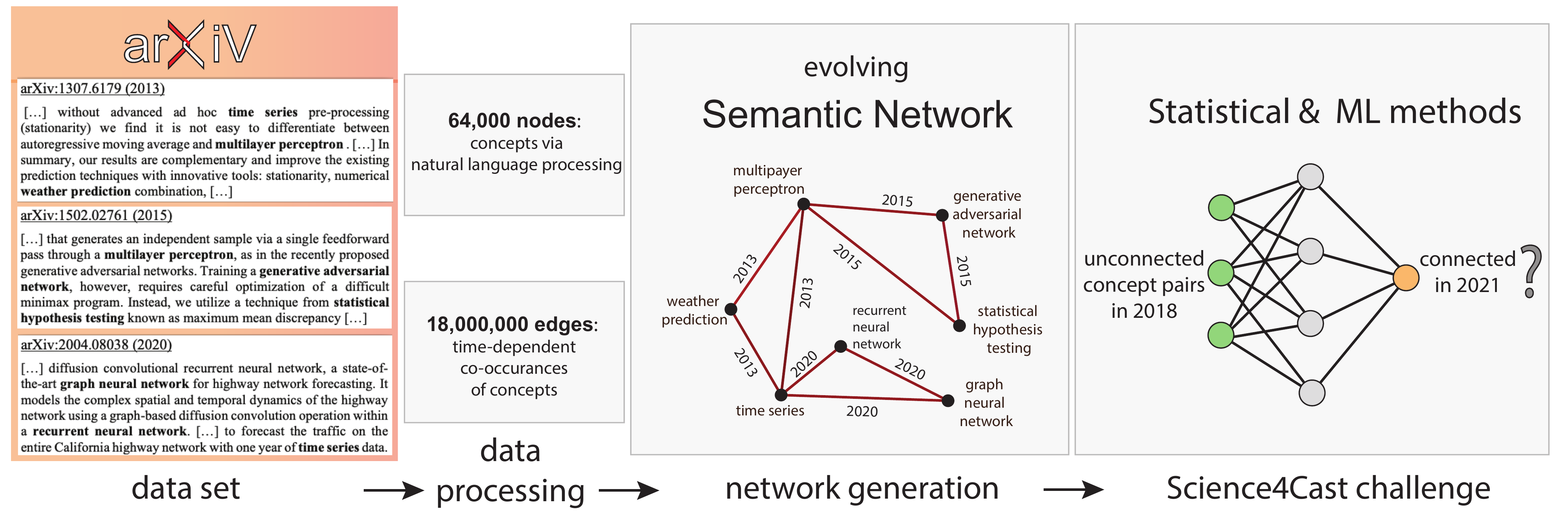}
\caption{\textbf{From arXiv to \texttt{Science4Cast}.} We use 143,000 papers in AI and ML categories on arXiv from 1992 to 2020. From there, we construct a list of concepts (using RAKE and other NLP tools). Those concepts form the nodes of a semantic network. The edges are drawn when two concepts occur jointly in the title or abstract of a paper. In that way, we generate an evolving semantic network that grows over time as more concepts are investigated together. The task is to predict, from unconnected nodes (i.e. concepts that have not been investigated together in the scientific literature), which will be connected within a few years. In this manuscript, we present 10 diverse statistical and machine learning methods to solve this challenge.} 
\label{fig:workflow}
\end{figure*}

\section{Semantic Networks}
The goal here is to extract knowledge from the scientific literature that can subsequently be processed by computer algorithms. At first glance, a natural first step would be to use the features of a large language model (such as GPT3 \cite{brown2020language}, Gopher \cite{rae2021scaling}, MegaTron \cite{smith2022using} or PaLM \cite{chowdhery2022palm}) from the text of each article to extract concepts automatically. However, those methods still struggle in reasoning capabilities \cite{kojima2022large, zhang2022paradox}, thus it is not yet directly clear how these models can be used for identifying and suggesting new ideas and concept combinations.

An alternative approach has been pioneered by Rzhetsky and colleagues \cite{rzhetsky2015choosing}. They have shown how knowledge networks (or semantic networks) in biochemistry can be created from co-occurring concepts in scientific papers. The nodes in their network correspond to scientific concepts---concretely, the names of individual biomolecules. The nodes are linked when a paper mentions both of the corresponding biomolecules in its title or abstract. Taking millions of papers into account leads to an evolving semantic network that captures the history of the field. Using supercomputer simulations, non-trivial statements about the collective behaviour of scientists can be extracted, which allows for the suggestions of alternative and more efficient research behaviour \cite{foster2015tradition}. Of course, by creating a semantic network from concept co-occurrences, only a tiny amount of knowledge is extracted from each paper. However, if this process is repeated for a large dataset of papers, the resulting network captures nontrivial and actionable content.

The idea to build up a semantic network of a scientific discipline was then applied and extended in the field of quantum physics \cite{krenn2020predicting}. There, the authors (including one of us) built a network of more than 6,000 quantum physics concepts. The authors formulate the task of predicting new research trends and connections for the first time as an ML task. The task was to identify which concept pairs, which have never been discussed jointly in the scientific literature, have a high probability to be investigated in the near future. This prediction task was phrased as one component for personalized suggestions of new research ideas.

\subsection{Link Prediction in Semantic Networks}
Here we formulate the predictions of future research topics as a link prediction task in an exponentially growing semantic network in the field of AI. Two nodes that do not share an edge have not been mentioned together in the title or abstract of an existing scientific paper. Here, the goal is to predict which unconnected nodes will be connected in the future---that is, determine which scientific concepts that have not been researched yet \textit{will} be jointly researched in the future.

Link prediction is a very common problem in computer science that can be solved with classical metrics and features as well as machine learning techniques. From the network theory side, several works have studied local motif-based methods \cite{liben2007link, albert2004conserved, Zhou:2009, kovacs2019network, Cannistraci:2018}, often based on path-counting, while other methods have studied more global features using linear optimization \cite{pech2019link}, global perturbations \cite{lu2015toward} and stochastic block models \cite{guimera2009missing}. Other machine-learning works have tried to optimize over a combination of hundred of predictors \cite{ghasemian2020stacking}. Further discussion on these methods is available in a recent review on link prediction \cite{zhou2021progresses}.

In \cite{krenn2020predicting}, this task was solved by computing 17 hand-crafted features of the evolving semantic network. In the Science4Cast competition, the goal was to find more precise methods for link-prediction tasks in semantic networks (a semantic network of AI that is 10 times larger than the one in \cite{krenn2020predicting}). Specifically, on the one hand, we would like to determine which features are useful; on the other hand, we would also like to know whether this task can be solved efficiently without hand-crafted features. Here, we present results for both questions.

\subsection{Potential for Idea Generation in Science}
The long-term goal of predictions and suggestions in semantic networks is to provide new ideas to individual researchers. In a way, we hope to build a creative artificial muse in science \cite{krenn2022scientific}. We can bias or constrain the model to give research topics that are related to the research interest of individual scientists, or a pair of scientists to suggest topics for collaborations in an interdisciplinary setting. Important future questions concern the discovery of impactful and surprising suggestions, and suggestions that give more context than two scientific concepts.

\section{Generation and Analysis of the Dataset}
\subsection{Dataset Construction}
We use papers that are published on arXiv in the categories \texttt{cs.AI}, \texttt{cs.LG}, \texttt{cs.NE}, and \texttt{stat.ML}, from 1992 to 2020, to create a dynamic semantic network. The nodes stand for computer science and in particular artificial intelligence concepts. We create the list of concepts from the title and abstracts of all of the 143,000 papers. We use Rapid Automatic Keyword Extraction (RAKE) to create candidate concepts \cite{rose2010automatic}, and normalize the list using standard NLP techniques and other self-created methods. Ultimately, this leads to a list of 64,719 concepts.

These concepts form the nodes of the semantic network. The edges are drawn when two concepts co-appear in a title or abstract of a paper. Each edge has a time stamp, which is the publication date of the paper in which the concepts co-appear. Multiple edges with different time-stamps between two concepts are very common, as concept pairs can co-appear in many papers with different publication dates. As edges have time stamps, the entire semantic network is evolving in time. The workflow is depicted in Fig. \ref{fig:workflow}.

\subsection{Network-Theoretical Analysis}

\begin{figure}[b]
    \centering
    \includegraphics[width=\linewidth]{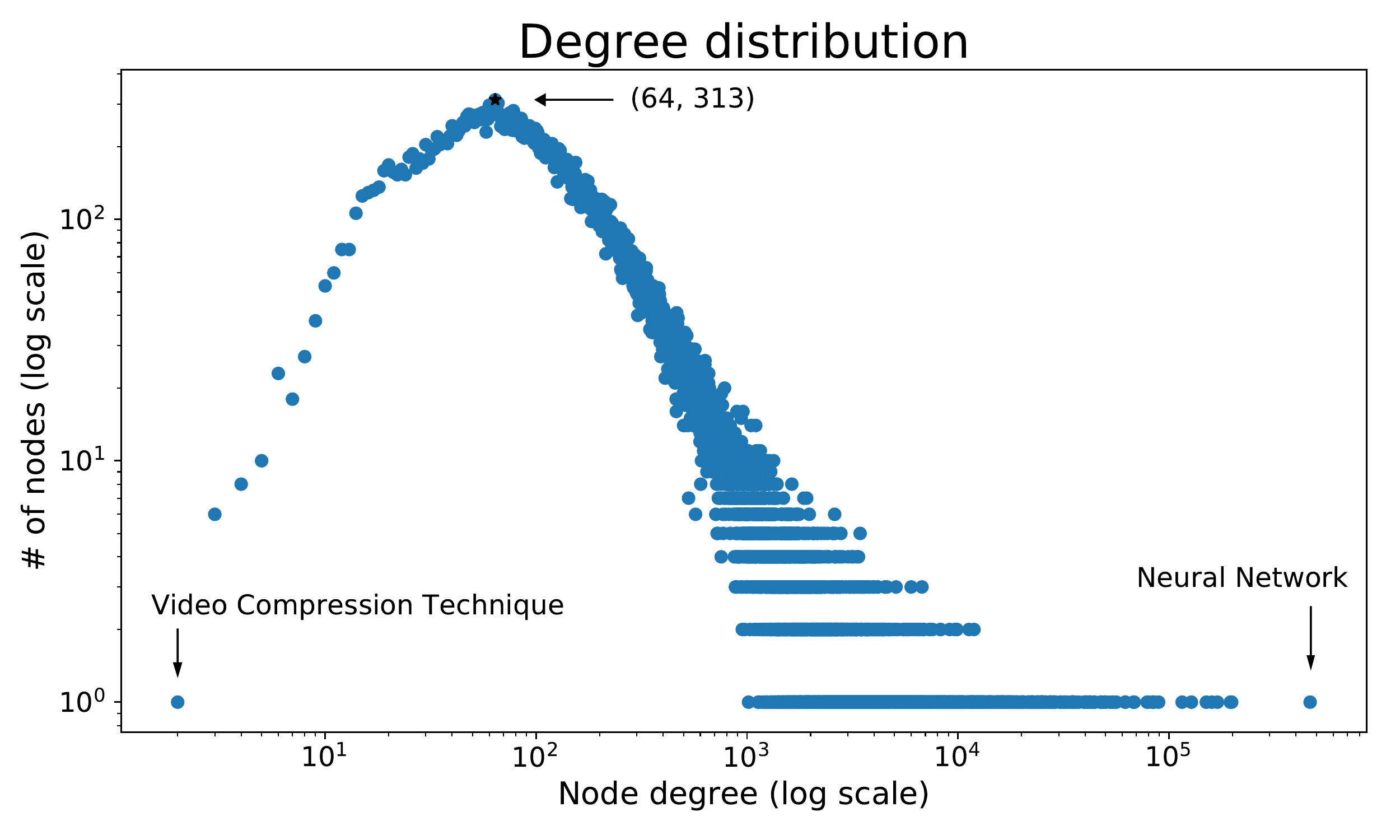}
    \caption{\textbf{Node degrees follow heavy-tail distribution due to the hubs.} Nodes with the largest (466,319) and smallest (2) non-zero  degrees correspond to \texttt{neural network} and \texttt{video compression technique}, respectively. The most common non-zero degree is 64. 1,247 nodes with zero degrees are not shown in this plot, and both axes are in log scale.}
    \label{fig:degree_histogram}
\end{figure}

We start by analyzing the degree distribution of the published semantic network. The network has 64,719 nodes and 17,892,352 unique undirected edges, which implies a mean node degree of about 553. However, the network contains many hub nodes that significantly exceed this mean degree, demonstrated by the heavy-tail degree distribution in Fig. \ref{fig:degree_histogram}. For example, the ten highest node degrees (and their corresponding concepts) are 466,319 (\texttt{neural network}), 198,050 (\texttt{deep learning}), 195,345 (\texttt{machine learning}), 169,555 (\texttt{convolutional neural network}), 159,403 (\texttt{real world}), 150,227 (\texttt{experimental result}), 127,642 (\texttt{deep neural network}), 115,334 (\texttt{large scale}),  89,267 (\texttt{high dimension}), and 84,956 (\texttt{high dimensional}).

\begin{figure}[htp]
    \centering
    \includegraphics[width=\linewidth]{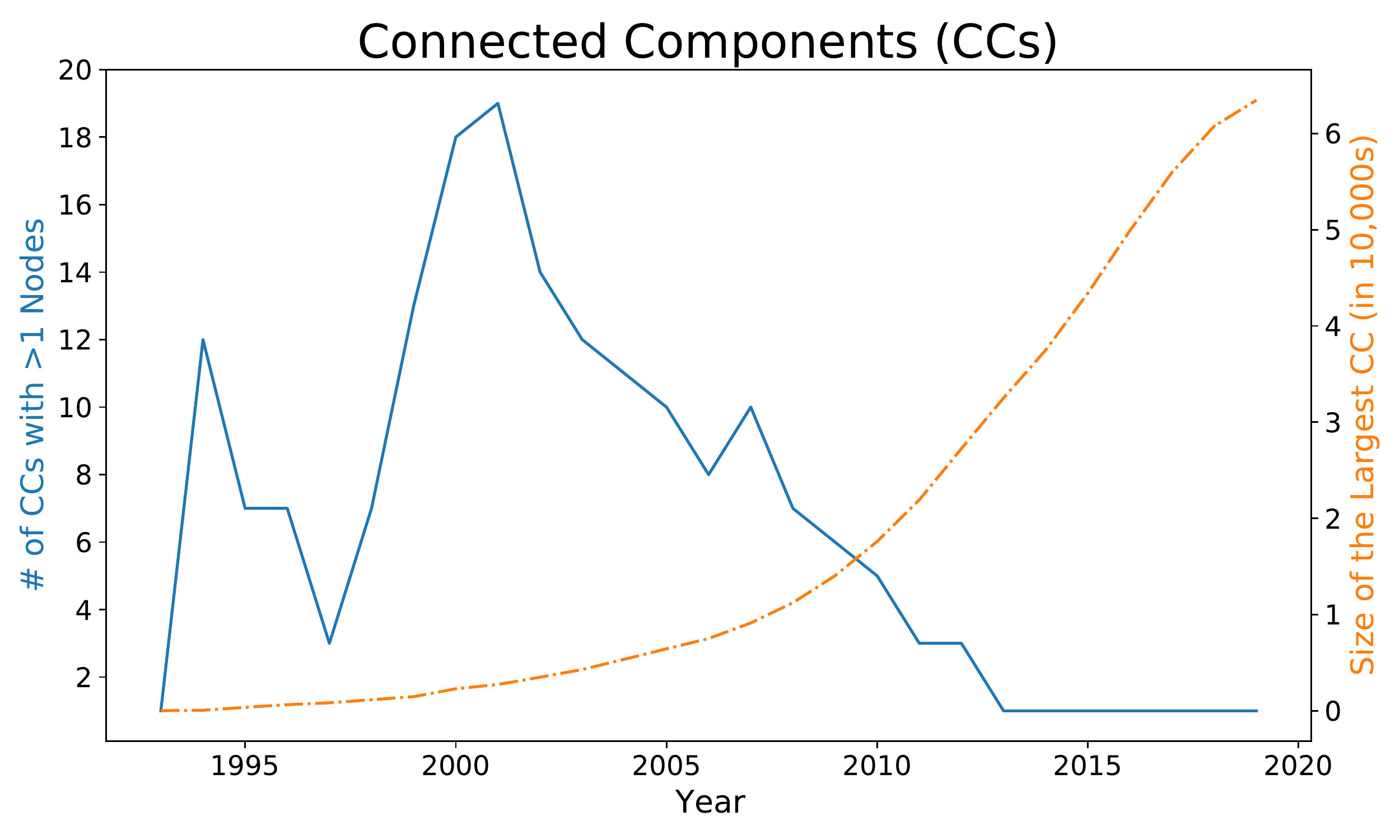}
    \caption{\textbf{The network became more connected over the years.} Primary (left, blue) vertical axis: Number of connected components with more than one node. Secondary (right, orange) vertical axis: Number of nodes in the largest connected component. For example, the network in 2019 comprises of one large connected component with 63,472 nodes and 1,247 isolated nodes, i.e. nodes with no edges. On the other hand, the 2001 network has 19 connected components with size greater than one, the largest of which has 2,733 nodes.}
    \label{fig:cc_numbers}
\end{figure}

To investigate whether this complex network is scale-free, we fit a power-law curve to the degree distribution $p(k)$ using \cite{alstott2014powerlaw}, and the software fit $p(k) \propto k^{-2.28}$ for degree $k \ge 1672$. Nevertheless, the degree distribution of real complex network do not always follow perfect power-laws and power-laws with exponential cut-offs are often a better fit than pure power-laws \cite{fenner2007model}.

A recent work \cite{broido2019scale} empirically showed that log-normal distributions fit most real-world networks as well as or better than power laws, and confirmed that pure ``scale-free networks are rare''. In light of that result, we used likelihood ratio tests to compare the power law fit with alternative distributions. The likelihood ratio tests from \cite{alstott2014powerlaw} suggested that truncated power law ($p$-value: 0.0031), lognormal ($p$-value: 0.0045), and lognormal positive ($p$-value: 0.015) fit the data better than power law, while exponential ($p$-value: 3e-10) and stretched exponential ($p$-value: 6e-05) were worse. We could not conclude whether truncated power law, lognormal, or lognormal positive \textit{best} describe the data with $p$-value $\le 0.1$.

Next, we discuss changes in the network connectivity over time. While the degree distributions maintained a heavy tail over the years, the ordering of the nodes inside the heavy tail changed, likely in response to the popularity trends in the field. The nodes with most connections (and the year they became so) are \texttt{decision tree} (1994), \texttt{machine learning} (1996), \texttt{logic program} (2000), \texttt{neural network} (2005), \texttt{experimental result} (2011), \texttt{machine learning} (2013), and finally, back to \texttt{neural network} (2015).

Furthermore, the network grew more connected over time according to connected component analysis in Fig. \ref{fig:cc_numbers}. Groups that were previously separated became connected, i.e. number of connected components decreased, while the largest group grew bigger. The trajectory of the mid-sized connected components may reveal interesting trends about their topics. Take image processing for instance. A connected component of the following 4 nodes appeared in 1999: \texttt{brightness change}, \texttt{planar curve}, \texttt{local feature}, and \texttt{differential invariant}. In 2000, 3 more nodes joined the group: \texttt{similarity transformation}, \texttt{template matching}, and
\texttt{invariant representation}. Then in 2006, a paper that discusses both \texttt{support vector machine} and	\texttt{local feature} merged this mid-size group of nodes to the largest connected component.

\begin{figure}[b]
\centering
\includegraphics[width=0.45\textwidth]{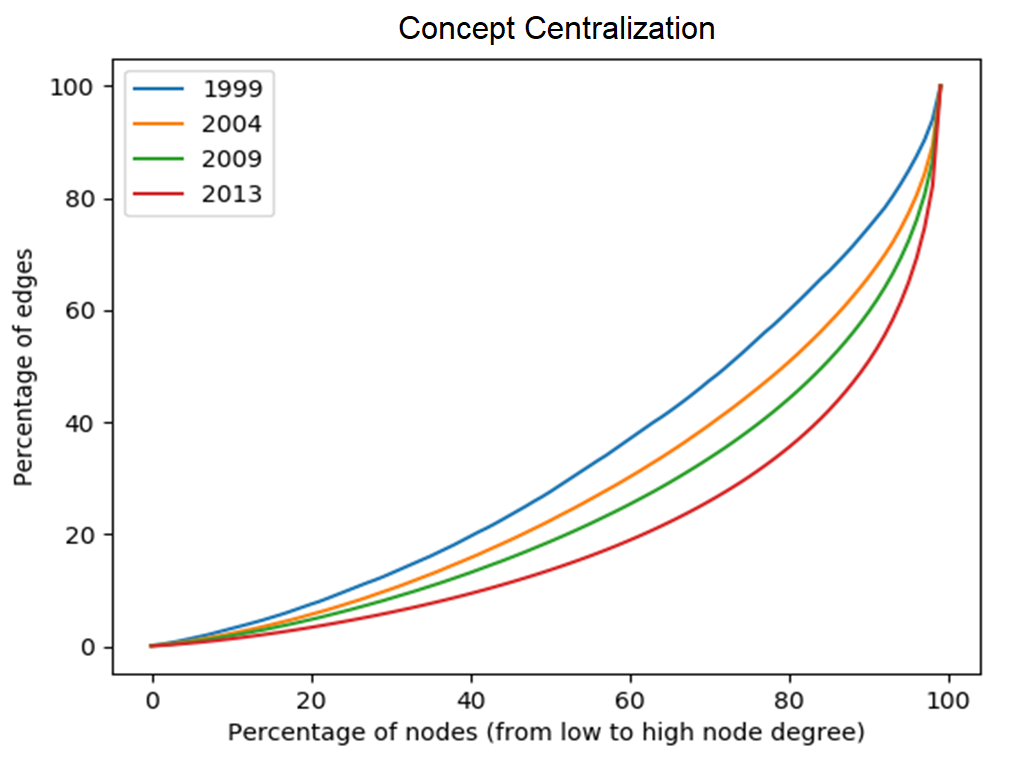}
\caption{\textbf{Centralization of Concepts.} The fraction of nodes (concepts) that corresponds to the fraction of edges (connections). Cumulative histogram of edges per node, up to a given year (1999,2004,2009,2014). The graph was created by going over the edges list and adding to each year only the edges and nodes that are dated before the year (hence the 2014 plot contains all the nodes (concepts) in papers before 2014). The nodes are arranged by increasing degrees. The plot is a  cumulative graph; hence the y value in the x=80, is the fraction of edges contributed by all the nodes in and below the 80s percentile of degrees.}
\label{fig:centralization}
\end{figure}

Another trend that emerges from the semantic network is an increase in centralization over time, with fewer percentage nodes (concepts) contributing larger fraction edges (concepts combination) over the years. This trend seems to be consistent across the entire period of the dataset. It can be seen from the histogram in Fig. \ref{fig:centralization} that the fraction of edges corresponding to the highest degree nodes (most connected) increases over the years, while the fraction of edges corresponding to the least connected nodes decreases. This trend is also consistent with the decrease in the average clustering coefficient over time (average clustering coefficient by year: 1999: \textit{0.919}, 2004: \textit{0.844}, 2009: \textit{0.773}, 2013: \textit{0.650}), implying most nodes are less likely to be connected with each other and more likely to be connected to a few high-degree central nodes. This trend might be explained by the fact that the AI community has been focusing on a few methods (e.g. deep learning) which have grown to dominate the field, compared to more diverse approaches in the 90s and 2000s. An alternative explanation is the use of more consistent terminology.

\begin{figure}[ht]
\centering
\includegraphics[width=0.45\textwidth]{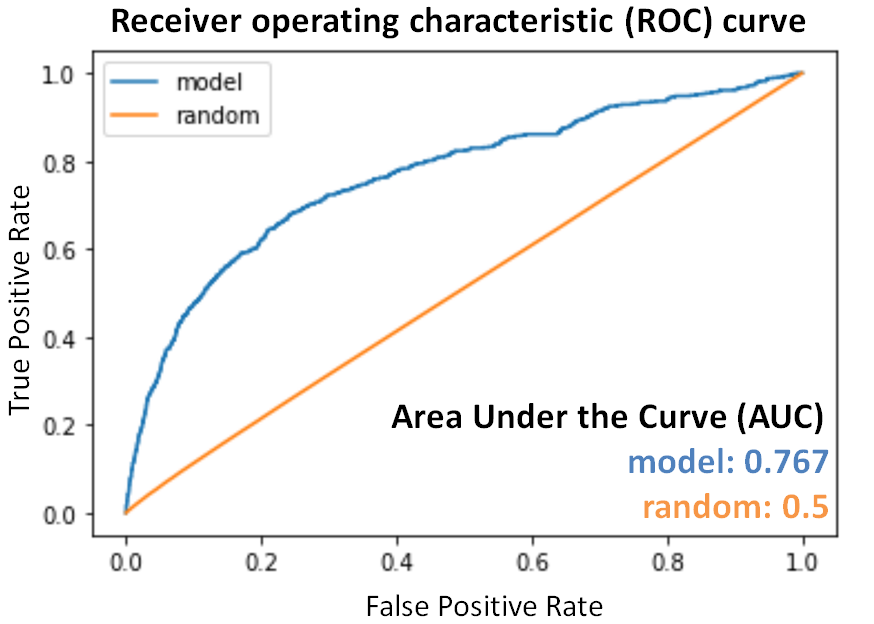}
\caption{\textbf{Receiver operating characteristic curve (ROC) for computing the Area under the Curve (AUC)}. Random Predictions get the result right in half of the cases, therefore their ROC curve is a diagonal with an AUC=0.5 (orange). A model that has learned some properties of the dataset has a $\text{AUC}>0.5$ (blue).}
\label{fig:ROC}
\end{figure}
\begin{figure*}[ht]
\centering
\includegraphics[width=0.7\textwidth]{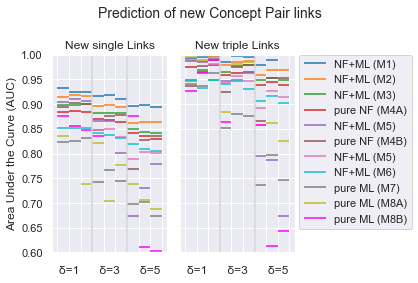}
\caption{\textbf{The \texttt{Science4Cast} benchmark: Link predictions in an exponentially growing semantic network.} Here we show the AUC values for different models that use machine learning techniques (ML), hand-crafted network features (NF) or a combination thereof. The left plot shows results for the prediction of a single new link (i.e., $w=1$), the right one shows results for the prediction of new triple links $w=3$. The task is to predict $\delta=[1,3,5]$ years into the future, with cutoff values $c=[0,5,25]$. We sort the models by the the results for the task ($w=1,\delta=3,c=0$), which was the task in the \texttt{Science4Cast} competition. Data points that are not shown have a AUC below 0.6 or are not computed due to computational costs. Note that the prediction of new triple edges can be performed nearly determinstically. It will be interesting to understand the origin of this quasi-deterministic pattern in AI research.}
\label{fig:AUCresults}
\end{figure*}
\subsection{Problem Formulation: Predictions in an exponentially growing semantic network}
The concrete task is to predict which two nodes $v_i$ with degrees $d(v_i)\leq c$ that do not share an edge in the year (2021-$\delta$) will have $w$ edges in the year 2021. We use $\delta=1,3,5$, $c=0,5,25$ and $w=1,3$. Note that $c=0$ is an interesting special case in which the node does not have any edge associated to it yet in the initial year. Thus, the model does not have any information about the node yet; the task there is to predict which nodes will be connected to entirely new edges. The task $w=3$ goes beyond simple link prediction, and asks which uninvestigated concept pair will be studied together in at least 3 papers.\footnote{One interesting alternative task is the prediction of the fastest growing links, which one could denote as \textit{trend} prediction.}

In the task, we provide a list of 10 million unconnected nodes pairs (each node having a degree$\leq$c) of the year (2021-$\delta$), and the goal is to sort this list from highest to lowest probability that in 2021 they will have at least $w$ edges. 

For the evaluation we use the ROC curve \cite{fawcett2004roc}; see Fig. \ref{fig:ROC} for details. The ROC curve is created by plotting the true positive rate against the false positive rate at various threshold settings. Our evaluation metric is the commonly used metric Area under the Curve (AUC) of the ROC curve. One advantage of AUC over mean-square-error is its independence of the data distribution. Specifically, in our case, where the two classes are highly asymmetrically distributed (with only about 1-3\% of newly connected edges), and the distribution changing over time, the AUC provides a meaningful and interpretation. For perfect predictions, AUC=$1$, while random predictions give AUC=$0.5$. It gives the percentage that a random true element is higher ranked than a random false one.

\section{AI-based Solutions}\label{solutions}
We now demonstrate how to solve this task with numerous different methods, from pure statistical approaches to hand-crafted features (NF) as an input of a neural network, to ML models that can work without hand-crafted features. All results are shown in Fig. \ref{fig:AUCresults}. The most powerful methods (those with the highest prediction quality measured by the AUC metric) take advantage of NF, which are the inputs to an ML model. Surprisingly, using purely network theoretical features without machine learning works competitively. Pure ML methods were not yet able to outperform those that use hand-crafted features. It remains an important open challenge how to solve this task without relying on hand-crafted features. While the prediction of new links can reach an AUC of up to 93\%, we find that the prediction of links that are generated at least three times can be solved with AUC>99.5\%. Understanding this apparently quasi-deterministic pattern in AI research will be an interesting target for follow-up research.\footnote{We have performed numerous additional tests to exclude data leakage in the benchmark dataset, overfitting or data duplication both in the set of articles and the set of concepts.}

\SkipTocEntry\subsection{M1: Features+ML} 
The solution of team oahciy is based on a blend of a tree-based gradient boosting approach and a graph neural network approach \cite{lu2021predicting}. Extensive feature engineering was conducted to capture the centralities of the nodes, the proximity between node pairs, and their evolution over time. The centrality of a node is captured by the number of neighbours and the PageRank score \cite{brin1998anatomy}, while the proximity between a node pair is derived using the Jaccard index. We refer the reader to \cite{lu2021predicting} for the list of all features and their feature importance.

The tree-based gradient boosting approach uses the Light Gradient Boosting Machine (LightGBM) \cite{ke2017lightgbm} and applies heavy regularization to combat overfitting due to the scarcity of positive samples. The graph neural network approach employs a time-aware graph neural network to learn node representations on dynamic semantic networks. 

\SkipTocEntry\subsection{M2: Features+ML} 
The method proposed by Team HashBrown assumes that the probability that nodes $u$ and $v$ form an edge in the future is a function of the node features $f(u)$, $f(v)$, and some edge feature $h(u,v)$. We chose node features $f$ that capture popularity at the current time $t_0$ (such as degree, clustering coefficient \cite{holland1971transitivity, watts1998collective}, and PageRank \cite{brin1998anatomy}). We also use these features' first and second time-derivatives to capture the evolution of the node's popularity over time. After variable selection during training, we chose $h$ to consist of the HOP-rec score \cite{yang2018hop, Lin2021} and a variation of the Dice similarity score \cite{sorensen1948method} as a measure of similarity between nodes. In summary, we use 31 node features for each node, and two edge features, which gives $31 \times 2 + 2 = 64$ features in total. These features are then fed into a small multilayer perceptron (MLP) (5 layers, each with 13 neurons) with ReLU activation.

Cold start is the problem that some nodes in the test set
do not appear in the training set. Our strategy for a cold start is imputation. We say a node $v$ is seen if it appeared in the training data, and unseen otherwise; similarly, we say that a node is born at time $t$ if $t$ is the first time stamp where an edge linking this node has appeared. The idea is that an unseen node is simply a node born in the future, so its features should look like a recently born node in the training set. If a node is unseen, then we impute its features as the average of the features of the nodes born recently. We found that with imputation during training, the test AUC scores across
all models consistently increased by about 0.02. For a complete description of this method, we refer the reader to \cite{tran2021improving}.

\SkipTocEntry\subsection{M3: Features+ML} 
This approach, detailed in \cite{sanjabi2021efficiently}, uses hand-crafted node features that have been captured in multiple time snapshots (e.g. every year) and then uses an LSTM to benefit from learning the time dependencies of these features. The final configuration uses two main types of features: node features including degree and degree of neighbours, and edge features including common neighbours. And to balance the training data the same number of positive and negative instances have been randomly sampled and combined.

One of the goals was to identify features that are very informative with a very low computational cost. We found that the degree centrality of the nodes is the most important feature, and the degree centrality of the neighbouring nodes and the degree of mutual neighbours gave us the best tradeoff. As all of the extracted features’ distributions are highly skewed to the right, meaning most of the features take near zero values, using a power transform like \textit{Yeo-Johnson} \cite{yeo2000new} helps to make the distributions more Gaussian which boosts the learning. Finally, for the link prediction task, we saw that LSTMs perform better than fully connected neural networks.

\SkipTocEntry\subsection{M4: pure Features } 
The following two methods are based on a purely statistical analysis of the test data and are explained in detail in \cite{moutinho2021network}.

\textbf{Preferential Attachment} -- In the network analysis we concluded that the growth of this dataset tends to maintain a heavy-tailed degree distribution, often associated with scale-free networks. As mentioned before the $\gamma$-value of the degree distribution is very close to $2$, suggesting that preferential-attachment \cite{barabasi2013network} is likely the main organizational principle of the network. As such, we implemented a simple prediction model following this procedure. Preferential-attachment scores in link prediction are often quantified as
\begin{equation}
    s_{ij}^\text{PA}=k_i\cdot k_j.
\end{equation}
with $k_{i,j}$ the degree of nodes $i$ and $j$. However, this assumes the scoring of links between nodes that are already connected to the network, that is $k_{i,j}>0$, which is not the case for all the links we must score in the dataset. 
As a result, we define our preferential attachment model as
\begin{equation}
    s_{ij}^\text{PA}=k_i + k_j. \label{eq:pa}
\end{equation}
Using this simple model with no free parameters we could score new links and compare them with the other models. Immediately we note that preferential attachment outperforms some learning-based models, even if it never manages to reach the top AUC, but it is extremely simple and with negligible computational cost.

\textbf{Common Neighbours} -- We explore another network-based approach to score the links. Indeed, while the preferential attachment model we derived performed well, it uses no information about the distance between $i$ and $j$, which is a popular feature used in link prediction methods \cite{zhou2021progresses}. As such we decided to test a method known as Common Neighbours \cite{liben2007link}. If we define $\Gamma(i)\cap\Gamma(j)$ as the set of common neighbours between nodes $i$ and $j$. We can easily score the nodes with
\begin{equation}
    s_{ij}^\text{CN}=|\Gamma(i)\cap\Gamma(j)| \label{eq:cn}
\end{equation}
the intuition being that nodes which share a larger number of neighbours are more likely to be connected than distant nodes that do not share any.

Evaluating this score for each pair $(i,j)$ on the dataset of unconnected pairs, which can be computed as the second power of the adjacency matrix, $A^2$, we obtained an AUC which is sometimes higher than preferential attachment and sometimes lower than it but is still consistently quite close with the best learning-based models.

\SkipTocEntry\subsection{M5: Features + ML} 

This method is based on \cite{francisco2021} with a modification disclosed in the \ref{appenix_m6}. First, 10 groups of first-order graph features are extracted to get some neighbourhood and similarity properties from each pair of nodes: degree centrality of nodes, pair's total number of neighbours, common neighbours index, Jaccard coefficient, Simpson coefficient, geometric coefficient, cosine coefficient, Adamic-Adar index, resource allocation index, and preferential attachment index. They are obtained for three consecutive years to capture the temporal dynamics of the semantic network, leading to a total of 33 features. Second, principal component analysis (PCA) \cite{pca2016} is applied to reduce the correlation between features, speed up the learning process and improve generalization, which results in a final set of 7 latent variables. Lastly, a random forest classifier is trained (using a balanced dataset) to estimate the likelihood of new links between the AI concepts.

\SkipTocEntry\subsection{M6: Features+ML} 
The baseline solution for the Science4Cast competition was closely related to the model presented in \cite{krenn2020predicting}. It uses 15 hand-crafted features of a pair of nodes $v_1$ and $v_2$ (Degrees of $v_1$ and $v_2$ in the current year and previous two years, these are six properties. The number of shared neighbours in total of $v_1$ and $v_2$ in the current year and previous two years are six properties. The number of shared neighbours between $v_1$ and $v_2$ in the current year and the previous two years, these are 3 properties). These 15 features are the input of a neural network with four layers (15, 100, 10, and 1 neurons), intending to predict whether the nodes $v_1$ and $v_2$ will have $w$ edges in the future. After the training, the model computes the probability for all 10 million evaluation examples. This list is sorted and the AUC is computed.

\SkipTocEntry\subsection{M7: end-to-end ML (Transformers)}

This model, which is detailed in  \cite{lee2021dynamic}, does not use any handcrafted features but learns them in a completely unsupervised manner. To do so, we extract various snapshots of the adjacency matrix through time, capturing graphs in the form of $\bm{A}_t$ for $t=1994, \ldots, 2019$. We then embed each of these graphs into 128-dimensional Euclidean space via Node2vec \cite{grover2016node2vec, pecanpy}. For each node $u$ in the semantic graph, we extract different 128-dimensional vector embeddings $\bm{n}_u(\bm{A}_{1994}), \ldots, \bm{n}_u(\bm{A}_{2019})$.

Transformers have performed extremely well in natural language processing tasks \cite{Vaswani2017AttentionIA}, thus we apply them to learn the dynamics of the embedding vectors. We pre-train a transformer to help classify node pairs. For the transformer, the encoder and decoder had 6 layers each; we used 128 as the embedding dimension, 2048 as the feedforward dimension and 8-headed attention. This transformer acts as our feature extractor. Once we pre-train our transformer, we add a 2-layer ReLU network with hidden dimension $128$ as a classifier on top.

\SkipTocEntry\subsection{M8: end-to-end ML (auto node embedding)} 

The most immediate way one can apply machine learning to this problem is by automating the detection of features. Quite simply, the baseline solution M6 is modified such that instead of 15 hand-crafted features, the neural network is instead trained on features extracted from a graph embedding. In our approach, we use the ProNE embedding \cite{ijcai2019-594}, which is based on sparse matrix factorizations modulated by the higher-order Cheeger inequality \cite{https://doi.org/10.48550/arxiv.1204.3873}, as well as Node2Vec \cite{grover2016node2vec}. We use the implementations provided in the \texttt{nodevectors} Python package \cite{Ranger2021}.

The embeddings learn a 32-dimensional representation for each node; hence, each edge representation is normalized to a single point in $[0, 1]^{64}$, and the concatenated features are the input of a neural network with two hidden layers of size 1000 and 30, respectively. Similarly to M6, the model is then tasked with computing the probability for the evaluation examples, which lets us determine the ROC.

\section{Extensions and Future Work}
Creating an AI that can suggest research topics to human scientists is highly ambitious and challenging. The present work of link prediction for a temporal network to draw connections between existing concepts is only the first step. We point out several extensions and future works that are directly relevant to the overarching goal of AI for AI.

\textbf{High-quality predictions without feature engineering} -- Surprisingly, given a graph with already extracted concepts as nodes and edges plotting the time evolution of joint appearance of these concepts in publications, the most powerful methods all used carefully hand-crafted features. It will be interesting to see whether end-to-end deep learning methods can solve tasks without feature engineering.

\textbf{Fully automated concept extraction} -- The concept list at the moment is created by a purely statistical text analysis using RAKE. The suggestions by RAKE are then manually inspected and phrases that do not correspond to a \textit{concept} are removed. While this process can be partially automated (as RAKE often makes the same mistakes which can be captured automatically), it is not a scalable process if one wants to create concept lists for the much larger corpus of science and engineering. A fully automated natural language processing algorithm that can extract meaningful concepts with minimal mistakes would be extremely useful.

\textbf{Generation of new concepts} -- Here we predict the emergence of links between two known concepts. One important question is whether an AI algorithm can compose words and generate new concepts. Different from the current work that is mostly supervised, the generation of new concepts is \textit{unsupervised}, hence more difficult. One approach to address this question has been presented in \cite{salatino2017topics,salatino2018augur}. There the authors can detect clusters of concepts with specific dynamics that indicate the formation of a new concept. It will be interesting to see how such emerging concepts can be incorporated into the current framework and used for suggestions for new research topics. 


\textbf{Semantic information beyond concept pairs} -- At the moment, every article's abstract and title are compressed into several links between concept pairs. This procedure does not represent all information in the article's abstract (let alone, the article itself). The more information one can extract from the article, the more meaningful the predictions and suggestions will be. Extending the representation of the semantic network to more complex data structures, such as hypergraphs \cite{battiston2021physics} are likely to be computationally more demanding but could significantly improve the prediction qualities. It might be also possible to find some ways to decrease the complexity of the analysis using clever tricks. For example, the authors in  \cite{PhysRevLett.124.248301} showed that the maximum node and hyperedge cover problem, two computational NP-hard problems, can be solved in polynomial time for most of the real-world hypergraphs tested. Whether such tricks exist for hyperlink prediction is still an open problem. The inclusion of sociological factors, such as the status of the involved researchers and their affiliations might help in prediction tasks. 

\textbf{Predictions of scientific success} -- The prediction of a new link between nodes in the semantic network means that we predict which concepts scientists will study for the first time in the future. This prediction however does not say anything about the potential importance and impact of the new connection. As a tool for high-quality suggestions, we need to introduce the prediction of a \textit{metric-of-success}, for example, estimated citation numbers of the new link or the rate of citation growth over time. This extension seems reasonable given that modelling and predictions of citation information in citation networks (where nodes are papers) is a prominent area of research within the science of science \cite{liu2019link, Reisz_2022}. Adapting these techniques to semantic networks will be an interesting future research direction.

\textbf{Anomaly detections} -- In a way, predicting the most likely new connection between concepts does not necessarily directly coincide with the goal of suggestions of new surprising research directions. After all, those links are predictable, thus potentially not surprising by themselves. While we believe that this type of prediction can still be a very useful contribution for \textit{suggestions}, there is another way to more directly find surprising combinations, namely by finding anomalies in the semantic networks. Those are potential links that have extreme properties in some metrics. There are powerful deep learning methods for anomaly detection \cite{kwon2019survey,pang2021deep} and their application in the semantic network presented here might be very interesting. In fact, while scientists tend to study topics in which they are already directly involved \cite{fortunato2018science,wang2021science}, often higher scientific impact results from the unexpected combination of more distant domains \cite{rzhetsky2015choosing}, which foster the search for those surprising and impactful associations.

\textbf{End-to-end formulation} -- As outlined above, we necessarily decomposed our goal of extracting knowledge from the scientific literature into two subtasks: extracting concepts and building and predicting the evolution of a semantic network resulting from those concepts. This stands in contrast to the dominant paradigm in deep learning that emerged over the last decade of so-called `end-to-end' training based on early spectacular successes \cite{collobert2011natural,krizhevsky2012imagenet,mnih2015human,silver2016mastering}. In this paradigm, problems are not broken into sub-problems but solved directly using deep differentiable architecture components trained via back-propagation \cite{glasmachers2017limits, lecun2015deep}. If such an `end-to-end' solution approach to our goal could be achieved it would be interesting to see whether it could replicate the success this deep learning paradigm had in other areas. 

\textbf{Human level machine comprehension} --
One of the defining goals of the Dartmouth Summer Research Project on Artificial Intelligence in 1956 was the following: `An attempt will be made to find how to make machines use language, form abstractions and concepts, solve kinds of problems now reserved for humans, and improve themselves.' \cite{mccarthy2006proposal}. Such an algorithm would be expected to handle an evolution in concept denotations due to new insights (i.e. the emergence of the term `Gibbs entropy' to distinguish Boltzmann's original concept of thermo-dynamical entropy as opposed to seeing it in the light of the more general emergent `Shannon entropy' or `von Neumann Entropy') or due to disputed originality (i.e. Bolai-Lobatchevskian Geometry and Hyperbolic Geometry are the same concept). An algorithm with such natural language understanding capabilities would thus be extremely useful to get closer to our goal. Although large language models and other multimodally trained language models like CLIP \cite{radford2021learning} or CLOOB \cite{furst2021cloob} have achieved outstanding results recently, it is an open question how much statistically trained natural language models alone could eventually form concepts and abstractions on a human level \cite{mitchell2021abstraction, lecun2022path}.   

\section{Conclusion}
Here we present a new AI benchmark for link prediction in exponentially growing semantic networks. Several of the solutions have been collected in the IEEE BigData Competition \textit{Science4Cast} in fall 2021, and generalized to the mode diverse tasks presented here. The goal was to boost the capabilities of predicting future research directions in the field of AI itself, which grows enormous over the decade. This ability might be an important part of a tool that gives personalized research suggestions to human scientists in the future. We find, rather surprisingly, that the prediction of strong new links (those that are formed three or more times) can be predicted with extremely high quality (AUC beyond 99\%). It will be interesting to investigate this quasi-deterministic pattern in AI research in more detail. The best methods used a clever combination of hand-crafted features and machine learning. It will be interesting whether pure learning methods, without hand-crafted features, will achieve high-quality results in the future. We also point out a number of open problems towards the goal of practical, personalized, interdisciplinary AI-based suggestions for new impactful research direction -- which we believe could become a disruptive tool in the future.

\SkipTocEntry\section*{Appendix}

\SkipTocEntry\subsection{Model availability}
All of the models described above can be found on GitHub.
\href{https://github.com/YichaoLu/Science4cast2021}{M1},
\href{https://github.com/princengoc/s4s-final}{M2}, 
\href{https://github.com/nimasnjb/sci4cast}{M3}, 
\href{https://github.com/Buffoni/quantum-link-prediction}{M4}, 
\href{https://github.com/pfranciscovalente/science4cast_topological}{M5}, 
\href{https://github.com/MarioKrenn6240/FutureOfAIviaAI}{M6}, 
\href{https://github.com/HarlinLee/science4cast}{M7}, 
\href{https://github.com/ag8/embedding-model-code}{M8}.

\SkipTocEntry\subsection{Details on M9}
The solution M9 was not part of the \texttt{Science4Cast} competition and therefore not described in the corresponding proceedings, thus we want to add more details. We compare the ProNE embedding to Node2Vec, which is also commonly used for graph embedding problems. The algorithm maps each node of the network to a point in 32-dimensional space based on a biased random walk procedure, which is fundamentally parameterized by two variables---$p$, the ``return parameter'', and $q$, the ``in-out parameter''. The return parameter determines the frequency of backtracking in the random walk, while the in-out parameter determines whether to bias the exploration to nearby nodes or distant nodes. Notably, these parameters significantly affect how the network is encoded---for instance, in the BlogCatalog dataset, optimal parameters were $p=0.25, q=0.25$, whereas for the Wikipedia graph, they were $p=4, q=0.5$ \cite{grover2016node2vec}. In initial experiments, we used the default $p=q=1$ for a 64-dimensional encoding, before feeding it into the same neural network as for the ProNE experiment. The higher variance in Node2Vec-based predictions likely has to do with the method's significant sensitivity to its hyperparameters. While ProNE is clearly better suited for a general multi-dataset link prediction problem, Node2Vec's parameter sensitivity may help us identify what features of the network are most important for predicting its temporal evolution.

\SkipTocEntry\subsection{Consideration for Model M6} \label{appenix_m6}

In this manuscript, a modification was performed in relation to the original formulation of the method \cite{francisco2021}: two of the original features, average neighbor degree and clustering coefficient, were infeasible to extract for some of the tasks covered in this paper, as their computation can be heavy for such a very large network, and they were discarded. Due to some computational memory issues, it was not possible to run the model for some of the tasks covered in this study, and so those results are missing.

\SkipTocEntry\section*{Acknowledgements}
The authors thank IARAI Vienna and IEEE for supporting and hosting the IEEE BigData Competition \textit{Science4Cast}. We are specifically grateful to David Kreil, Moritz Neun, Christian Eichenberger, Markus Spanring, Henry Martin, Dirk Geschke, Daniel Springer, Pedro Herruzo, Marvin McCutchan, Alina Mihai, Toma Furdui, Gabi Fratica, Miriam V\'{a}zquez, Aleksandra Gruca, Johannes Brandstetter and Sepp Hochreiter for helping to set up and successfully execute the competition and the corresponding workshop. The authors thank Xuemei Gu for creating Fig.\ref{fig:workflow}, and Milad Aghajohari and Mohammad Sadegh Akhondzadeh for helpful comments on the manuscript. The work of HL, RS, and JGF were supported by grant TWCF0333 from the Templeton World Charity Foundation. HL is additionally supported by NSF grant DMS-1952339. JPM acknowledges the support of FCT (Portugal) through scholarship SFRH/BD/144151/2019. BC thanks the support from FCT/MCTES through national funds and when applicable co-funded EU funds under the project UIDB/50008/2020, and FCT through the project CEECINST/00117/2018/CP1495/CT0001. NMT and YX are supported by NSF Grant DMS-2113468, the NSF IFML 2019844 award to the University of Texas at Austin, and the Good Systems Research Initiative, part of University of Texas at Austin Bridging Barriers.

\bibliography{refs}
\end{document}